\pdfoutput=1

\documentclass[11pt]{article}

\usepackage[final]{acl}

\usepackage{times}
\usepackage{latexsym}

\usepackage[T1]{fontenc}

\usepackage[utf8]{inputenc}

\usepackage{microtype}

\usepackage{fancyhdr}
\usepackage{color}
\usepackage{url}
\usepackage{multirow}
\usepackage{graphicx} %
\usepackage{times}
\usepackage{latexsym}
\usepackage{multicol}
\usepackage{multirow}


\graphicspath{{fig/}} %

\usepackage{placeins}
\usepackage{graphicx}
\usepackage{balance}
\usepackage{soul}
\usepackage[ruled,vlined]{algorithm2e}

\definecolor{clabelbox}{rgb}{0.25, 0.25, 0.25}

\usepackage{bbm}
\usepackage{bm}
\usepackage{booktabs}

\usepackage{graphicx}
\graphicspath{{../figures/}{figures/}}

\newcommand{\fix}[1]{}

\newcommand{\ignore}[1]{}

\newtheorem{definition}{Definition}
\newcommand{\methodFull}{\textsc{mask neuron coverage}\xspace}
\newcommand{\methodName}{\textsc{mnCover}\xspace}
\newcommand{\method}{\textsc{mnCover}\xspace}
\newcommand{\methodNoMask}{\textsc{Cover}\xspace}

\newcommand{\imask}{\textsc{WIMask}\xspace}
\newcommand{\CheckList}{\textsc{CheckList}\xspace}

\newcommand{\Checklist}{\textsc{CheckList}\xspace}

\newcommand{\layer}{\imask layer}

\usepackage[nolist,nohyperlinks]{acronym}

\makeatletter
\newcommand{\removelatexerror}{\let\@latex@error\@gobble}
\makeatother

\usepackage{lipsum}
\usepackage{titlesec}

\titlespacing\subsection{0pt}{3pt plus 1pt minus 1pt}{1pt plus 0pt minus 0pt}
\titlespacing\subsubsection{0pt}{1pt plus 0pt minus 1pt}{0pt plus 0pt minus 0pt}
\titlespacing{\paragraph}{0pt}{2pt}{1pt}[0pt]

\usepackage{etoolbox}
\makeatletter
\preto{\@tabular}{\parskip=3pt}
\makeatother

\usepackage{enumitem}
\setlist[itemize]{leftmargin=*}
\setlist[itemize]{noitemsep, topsep=0pt}

\usepackage{tabularx} %
\usepackage{amssymb}%
\usepackage{amsmath}
\usepackage{mathtools}
\usepackage{pifont}%

\usepackage{ragged2e}

\usepackage{hyperref}
\usepackage{url}
\usepackage{dirtytalk}

\newcommand{\tref}[1]{Table~\ref{#1}} 

\newcommand{\sref}[1]{Section~(\ref{#1})}

\newcommand{\cref}[1]{Condition~(\ref{#1})} 
\newcommand{\dref}[1]{Definition~(\ref{#1})}

\include{macros}
\def\N{\mathbb{N}}
\def\E{\mathbf{E}}
\def\e{{\mathbf e}} 
\def\x{{\mathbf x}} 
\def\z{{\mathbf z}} 
\def\y{{\mathbf y}} 
\def\X{{\mathbf X}} 
\def\bZ{{\mathbf Z}}

\def\h{{\mathbf h}} 
 
\def\W{{\mathbf W}}

\def\data{{\mathbf D}} 
\def\neu{{\mathbf n}} 
\def\attmask{\mathbf{M}_{A}}
\def\wmask{\mathbf{M}_{W}}
\def\R{\wmask}
\def\inx{\mathbf{x}}
\def\mR{\wmask}
\def\mA{{\attmask}}

\def\x{{\mathbf x}} 
\def\z{{\mathbf z}} 
\def\y{{\mathbf y}} 
 
\def\h{{\mathbf h}} 
 
\def\W{{\mathbf W}}

\usepackage{colortbl}

\usepackage{booktabs}

\usepackage[nolist,nohyperlinks]{acronym}

\usepackage{lipsum}
\usepackage{titlesec}

\usepackage{etoolbox}
\makeatletter
\preto{\@tabular}{\parskip=5pt}
\makeatother

\usepackage{enumitem}
\setlist[itemize]{leftmargin=*}
 \setlist[itemize]{noitemsep}
 
\usepackage[font=small,skip=3pt]{caption}

\usepackage{trackchanges}
\addeditor{YJ}

\title{White-box Testing of NLP models with Mask Neuron Coverage}
\author{Arshdeep Sekhon  \and Yangfeng Ji  \and Matthew B. Dwyer \and Yanjun Qi \\ 
        University of Virginia, USA}

\begin{document}
\maketitle

\begin{abstract}

Recent literature has seen growing interest in using black-box strategies like \CheckList for testing the behavior of NLP models. 
Research on white-box testing has developed a number of methods for evaluating
how thoroughly the internal behavior of deep models is tested, but they are not applicable
to NLP models.
We propose a set of white-box testing methods that are customized for transformer-based NLP models.
These include \methodFull (\method) that measures how thoroughly
the attention layers in models are exercised during testing.
We show that \method can refine testing suites generated by \CheckList by substantially
reduce them in size, for more than 60\% on average, while retaining failing tests -- thereby concentrating the fault
detection power of the test suite.
Further we show how \method can be used to guide \CheckList input generation,
evaluate alternative NLP testing methods, and drive data augmentation to improve accuracy.

\end{abstract}

\section{Introduction}

Previous NLP methods have used black-box testing to discover errors in  NLP models. For instance, Checklist\cite{ribeiro2020accuracy} introduces a black-box testing strategy  as a new evaluation methodology 
for comprehensive behavioral testing of NLP models. CheckList introduced different test types, such as prediction
invariance in the presence of certain perturbations.

Black-box testing approaches, like Checklist, may produce distinct test inputs that yield very similar internal behavior from an NLP model. Requiring that generated tests are distinct both from a black-box and a white-box perspective -- that measures test similarity in terms of latent representations -- has the potential to reduce the cost of testing without reducing its error-detection effectiveness.
Researchers have explored a range of white-box coverage techniques that focus on neuron activations and demonstrated their benefit on architecturally simple feed-forward networks~\cite{pei2017deepxplore, tian2018deeptest,ma2018deepgauge,Dolaetal:ICSE:2021}.  However, transformer-based NLP models incorporate more complex layer types, such as those computing self-attention, to which prior work is inapplicable.

In this paper, we propose a suite of white-box coverage metrics. We first adapt the k-multisection neuron coverage measure from \cite{ma2018deepgauge} to Transformer architectures. Then we design a novel \methodName coverage metric, tailored to NLP models. \methodName focuses on the neural modules that are important for NLP and designs strategies to ensure that those modules' behavior is thoroughly exercised by a test set. Our proposed coverage metric, when used to guide test generation, can cost-effectively achieve high-levels of coverage.

\begin{figure}
    \centering
    \includegraphics[width=\columnwidth]{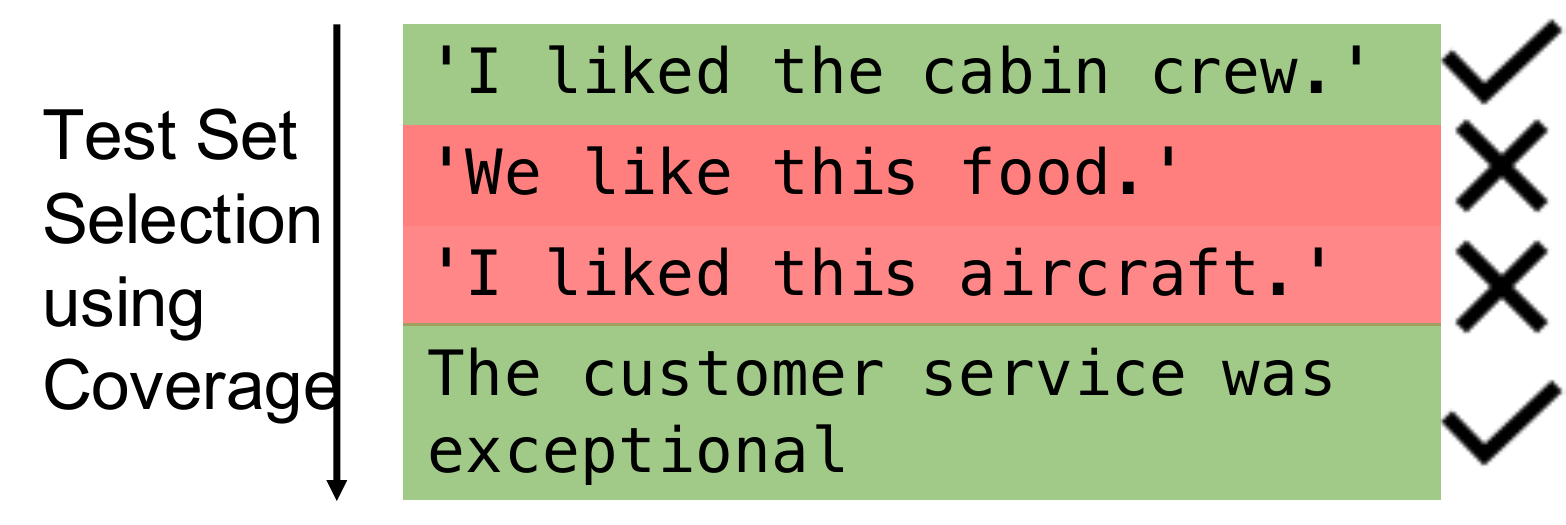}
    \caption{Example of our proposed \method's filtering on a set of test examples.}
    \label{fig:cover_example}
\end{figure}

Figure~\ref{fig:cover_example} shows one example of how \methodName can work in concert with CheckList to produce a small and effective test set. The primary insight is that not all text sentences contain new information that will improve our confidence in a target model's behavior. In this list, multiple sentences were generated with similar syntactic and semantic structure. These sentences cause the activation of sets of attention neurons that have substantial
overlap. This represents a form of redundancy in testing an NLP model.
Coverage-based filtering seeks to identify when an input's activation of attention neurons
is subsumed by that of prior test inputs -- such inputs are filtered.
In the Figure the second and third sentences are filtered out because their activation
of attention neurons is identical to the first test sentence.   As we show in 
\S\ref{sec:experiments} this form of filtering can substantially reduce test suite
size while retaining tests that expose failures in modern NLP models, such as BERT.

The primary contributions of the paper lie in:
\begin{itemize}
\item Introducing \method a test coverage metric designed to address the attention-layers that are characteristic of NLP models and  to account for the data distribution by considering task-specific important words and combinations.
\item Demonstrating through experiments on 2 NLP models (BERT, Roberta), 2 datasets (SST-2, QQP), and 24 sentence transformations that \method can substantially reduce the size of test sets generated by CheckList, by 64\% on average,  while improving the failure detection of the resulting tests, by 13\% on average.
\item Demonstrating that \method provide an effective supplementary criterion for evaluating the quality of test sets and that it can be used to generate augmented training data that improves model accuracy.
\end{itemize}

\section{Background}

\paragraph{Coverage for testing deep networks} The research of Coverage testing focuses on the concept of "adequacy criterion" that defines when ``enough'' testing has been performed. 
The white-box coverage testing
 has been proposed  
 by multiple recent studies  to test deep neural networks \cite{pei2017deepxplore,ma2018deepgauge,ma2018combinatorial,Dolaetal:ICSE:2021}. 
DeepXplore \cite{pei2017deepxplore}, a white-box differential testing algorithm,
introduced Neuron Coverage for DNNs to guide systematic exploration of DNN’s internal
logic. Let us use $\data$ to denote a set of test
inputs (normally named as a test suite in behavior testing). The Neuron Coverage regarding $\data$  is defined as the ratio between the number of unique activated neurons (activated by $\data$) and  the total number of neurons in that DNN under behavior testing.
A neuron is considered to be activated if its output is higher
than a threshold value (e.g., 0). Another 
closely related study,  DeepTest \cite{tian2018deeptest}, proposed a gray-box, neuron coverage-guided test suite generation strategy. Then, the study DeepGauge \cite{ma2018deepgauge} expands the neuron coverage definition by introducing  the kmultisection neuron coverage criteria to produce a multi-granular set of DNN coverage metrics. For a given  neuron $\neu$, the kmultisection neuron coverage measures how thoroughly a given
set of test inputs like $\data$ covers the range $[low_n, high_n
]$. %
The range $[low_n, high_n]$ is divided into k equal bins (i.e.,
k-multisections), for $k > 0$. For  $D$ and the target neuron
$\neu$, its k-multisection neuron coverage is then defined as the ratio of the
number of bins covered by $D$ and the total number of bins,
i.e., k. For an entire DNN model, the k-multisection neuron coverage is then the ratio of all the activated bins for all its neurons and the total number of bins for all neurons in the DNN.

\paragraph{Transformer architecture}
NLP is undergoing a paradigm shift with the rise of large scale Transformer models (e.g., BERT, DALL-E, GPT-3) that are
trained on unprecedented data scale and are adaptable to a wide range of downstream tasks\cite{bommasani2021opportunities}.

 These models  embrace the Transformer architecture \cite{vaswani2017attention} and 
can capture long-range 
pairwise or higher-order interactions between input elements. They utilize the self-attention mechanism\cite{vaswani2017attention}
that enables shorter computation paths and provides parallelizable computation for learning to represent a sequential input data, like text. 
Transformer receives inputs in the general form of word tokens.
The sequence of inputs is  converted to vector embeddings that  get repeatedly re-encoded  via the self-attention mechanism. The self-attention  can repeat for many layers, with each layer re-encoding and each layer maintaining the same sequence length.  At each layer, it corresponds to the following operations to learn encoding of token at position $i$: 
\begin{gather}
    \alpha^{ij} =
    \textrm{softmax}\big((\mathbf{W}^q\bm{h}_i)^{\top} (\mathbf{W}^k\bm{h}_j)/\sqrt{\smash[b]d}\big) \\
    \bm{\bar{h}}_{i}= \sum_{j=1}^{M} \alpha^{i j} \mathbf{W}^v \bm{h}_{j} \\
    \bm{h}'_{i} =  {\sigma}(\bm{\bar{h}}_{i}\mathbf{W}^r + \bm{b}_1)\mathbf{W}^o + \bm{b}_2.
    \label{eq:ffn}
\end{gather}
Here $\mathbf{W}^k$ is the key weight matrix, $\mathbf{W}^q$ is the query weight matrix,$\mathbf{W}^v$ is the value weight matrix, $\mathbf{W}^r$ and $\mathbf{W}^o$ are transformation matrices, and $\bm{b}_1$ and $\bm{b}_2$ are bias vectors.

\section{Method}

\begin{figure*}[t]
\begin{minipage}[]{\hsize}
   \centering
  \includegraphics[width=\columnwidth]{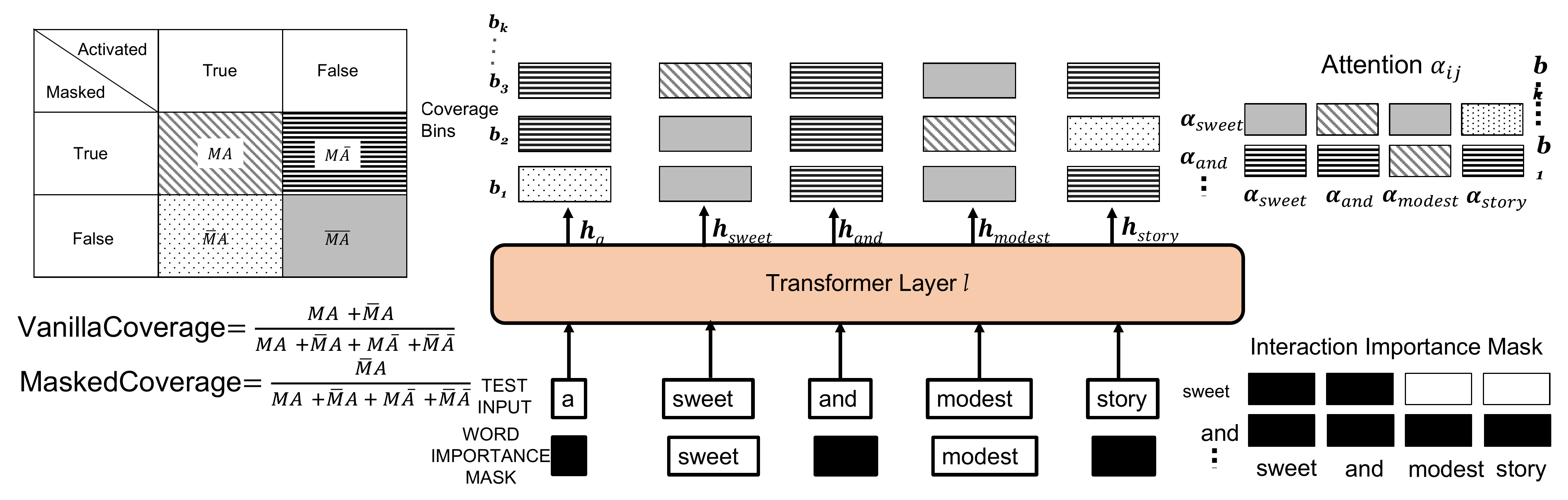}
   \caption{A visual depiction of \method for masking neurons to measure coverage.
   \label{fig:masking_coverage}}
\end{minipage}
\end{figure*}

State-of-the-art NLP models are large-scale with millions of neurons, due to large hidden sizes and multiple layers. We propose to simplify and view these foundation models~\cite{bommasani2021opportunities} through two levels of granularity: (1) {\it Word Level}: that includes the position-level embeddings at each layer and (2) {\it Pairwise Attention Level}: that includes the pairwise self-attention neurons between two positions at each layer.   %
In the rest of this paper, we denote the vector embeddings at location $i$ for layer $l$ as ${\h'}_{i}^l$ and name these as the {\it word level neurons} at layer $l$. We also denote  the $\alpha^{ij}$ at layer $l$ and head $h$ as $\alpha^{ij}_{lk}$, and  call them as the {\it attention level neurons} at layer $l$.

\subsection{Extending Neuron Coverage (\methodNoMask) for Testing NLP Model}
Now we use the above two layers' view we proposed, to adapt the vanilla neuron coverage concepts from the literature to NLP models. First, we introduce a basic definition: "activated neuron bins" \cite{ma2018combinatorial}: 
\begin{definition}
{Activated Neuron Bins (ANB): } For each neuron, we partition
the range of its values (obtained from training data)   into $B$ bins/sections. We define ANB for a given text input if the input's activation value from the target neuron  falls into the corresponding bin range. 

\end{definition}

Then we adapt the above definition to the NLP model setting, by using the after-mentioned two layers' view.  We design two phrases: Word Neuron Bins, and Activated Word Neuron Bins in the following ~\dref{def:awb}.
\begin{definition}
{ Activated Word Neuron Bins(AWB)}: We discretize the possible values of each neuron in $\mathbf{h}'^{\ell}_{t}$ (whose $d$-th embedding dimension is $\mathbf{h}'^{\ell}_{dt}$) into $B$ sections. We propose a function $\phi_w$ who takes two arguments, as   $\mathbf{\phi}_w(h^{'lb}_{dt}, \mathbf{x})$ for a given input $\inx$. $\mathbf{\phi}_w(h^{'lb}_{dt}, \mathbf{x})=1$ if it is an activated word neuron bin (shortened as AWB), else $0$ if not activated.
\label{def:awb}
\end{definition}

Similarly, for our attention neuron at layer $l$, head $k$, word position $i$ and position $j$: $\alpha_{ij}^{lk}$, we introduce the definition of "attention neuron bins" and "Activated Attention Neuron Bins" in the following \dref{def:aab}.
\begin{definition}
{ Activated Attention Neuron Bins (AAB)}: We discretize the possible values of neuron $\alpha_{ij}^{lk}$ into $B$ sections. We denote the state of the $b^{th}$ section of this attention neuron using $\mathbf{\phi}_a(\alpha^{ijb}_{lk}, \mathbf{x})$. $\mathbf{\phi}_a(\alpha^{ijb}_{lk}, \mathbf{x})=1$ if it is an activated attention neuron bin (denoted by AAB) by an input $\mathbf{x}$ and  $\mathbf{\phi}_a(\alpha^{ijb}_{lk}, \mathbf{x})=0 $ if not activated.
\label{def:aab}
\end{definition}

\begin{equation}
   \mathbb{N}(\text{\text{AWB}}(\mathbf{x})) = \sum_{ltdb} {\phi}_w(h^{'lb}_{dt}, \mathbf{x})
\end{equation}

\begin{equation}
    \mathbb{N}(\text{AAB}(\x)) = \sum_{ijbk} \mathbf{\phi}_a(\alpha^{ijb}_{lk}, \mathbf{x})
\end{equation}
The coverage, denoted by \methodNoMask, of a dataset $\mathcal{T}$ for a target model is then defined as the ratio between the number of ``activated" neurons and total neurons:
\begin{equation}
\methodNoMask = \dfrac{\mathbb{N}(\text{AWB}) + \lambda \mathbb{N}(\text{AAB})}{\mathbb{N}(\text{WB}) + \lambda \mathbb{N}(\text{AB})}  \label{eq:vanilla-coverage}
\end{equation}
Here, $\lambda$ is a scaling factor.

Now let us assume the total number of layers be $D$, total number of heads $H$, maximum length $L$, total bins $B$ and total embedding size be $E$.
 Considering the example case of the BERT\cite{devlin2019bert} model, total number of word level neurons to be measured are then $L\times E \times D = 128\times 768\times 13 \sim 0.1 million$. The total number of the attention level neurons is then $L\times L\times H\times D = 128\times 128\times 12 \times 12 \sim 2 million$.

\subsection{\methodFull (\method)}

 However, accounting for every word and attention neuron's behavior for a large pre-trained model like BERT is difficult for two reasons: (1). If we desire to test each neuron at the output of all transformer layers in each BERT layer, we need to account for the behavior of every neuron, which for a large pre-trained model like BERT is in the order of millions.
  
 (2). If we test every possible neuron, we need to track many neurons  that are  almost irrelevant for a target task and/or model. This type of redundancy makes the behavior testing less confident and much more expensive.

To mitigate these concerns, we propose to only focus on important words and their combinations that may potentially contain `surprising' new information for the model and hence need to be tested.

We assume we have access to a word level importance mask, denoted by $\mR \in \{0,1\}^{|V|}$ and the interaction importance mask by $\mA \in \{0,1\}^{|V| \times |V|}$.   Each entry in ${\mR}_{w_t} \in \{0,1\}$ represents the importance of word $w_t$. Similarly, each entry in $\mA_{{\x_i, \x_j}} \in \{0,1\}$ represents the importance of interaction between token $w_{t_i}$ and $w_{t_j}$. 
These masks aim for filtering out unimportant tokens (and their corresponding neurons at each layer) for measuring coverage signals. 
We apply the two masks at each layer to mask out unimportant attention pairs to prevent them from being counted towards coverage calculation. 
With the masks, the AWB and AAB are revised and we then define \methodFull (\method) accordingly: 
\begin{align}
    \vspace{-8mm}
& \N(\text{\text{Mask-AWB}}(\mathbf{x})) =
    \sum_{ltdb}{\mathbb{M}_{w}}_{\x_{t}}*{\phi}_w(h^{'lb}_{dt}, \mathbf{x}) \nonumber \\
& \N(\text{\text{Mask-AAB}}(\mathbf{x}))  = \sum_{ijkb}
    {\mathbb{M}_{a}}_{\x_{i}, \x_{j}}*\mathbf{\phi}_a(\alpha^{ijb}_{lk}, \mathbf{x}) \nonumber \\
& \method = \dfrac{\N(\text{Mask-AWB}) + \lambda \N(\text{Mask-AAB})}{\N(\text{WB}) + \lambda \N(\text{AB})}  \label{eq:mask-coverage-ii}
    \vspace{-5mm}
\end{align}
\vspace{-6mm}

\begin{figure}[t]
\begin{minipage}[]{\hsize}
    \centering
    \includegraphics[width=\hsize]{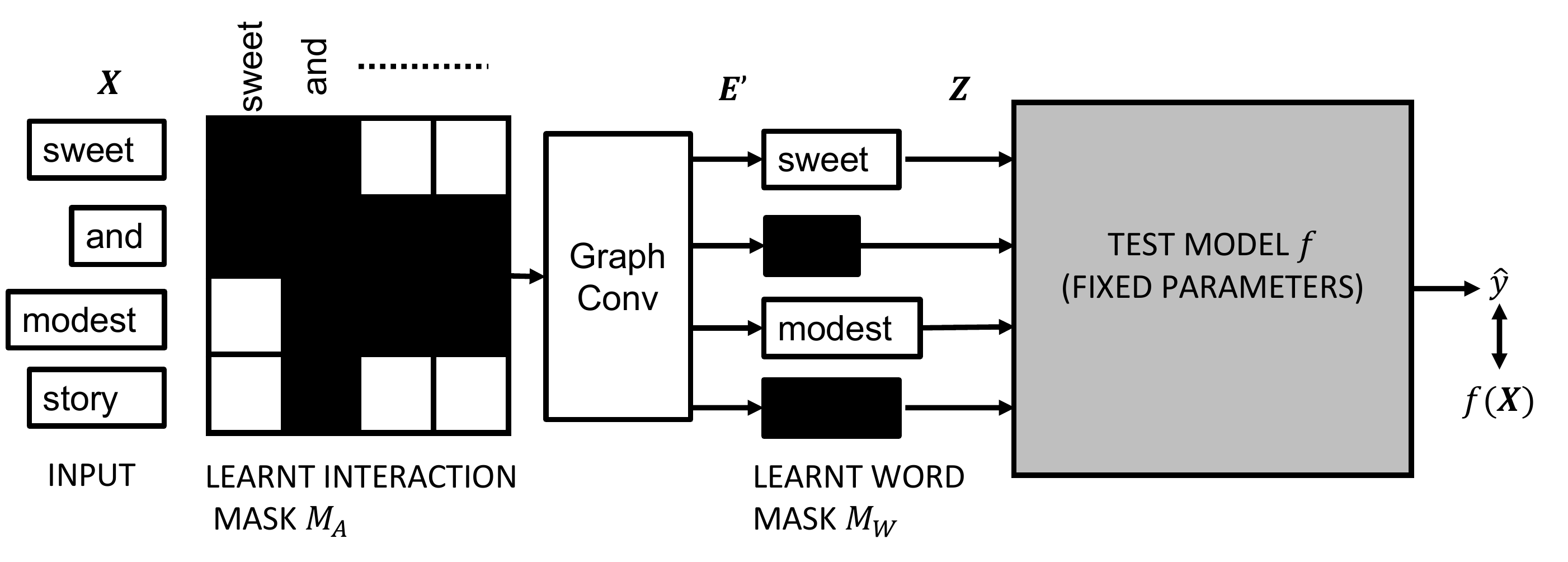}
    \caption{Learning the masks prior to testing: Globally important words and interactions are learnt by masking inputs to a target model.
    \label{fig:mask_to_model_2}}
\end{minipage}
\end{figure}

\subsection{Learning Importance Masks}

In this section, we explain our mask learning strategy that enables us to learn globally important words and their pairwise combinations for a model's prediction without modifying a target model's parameters.

We learn the two masks through a bottleneck strategy, that we call \imask layer. Given a target model $f$, we insert this mask bottleneck layer between  the word embedding layer  of a pretrained NLP model and the rest layers of this model.  
Figure~\ref{fig:mask_to_model_2} shows a high level overview of the mask bottleneck layer. Using our information bottleneck layer, we learn two masks : (1) a word level mask $\mA$, (2) an interaction mask $\mR$.

\paragraph{Learning Word-Pair Interaction Importance Mask: $\mA$}

The interaction mask  aims to discover which words globally interact for a prediction task. We treat words as nodes and represent their interactions as edges in an interaction graph.  We represent this unknown graph as a matrix $\attmask = {\{ {\attmask}_{ij} \}}_{V \times V}$. Each entry ${\attmask}_{\x_i, \x_j} \in \{0,1\}$ is a binary random variable, such that ${\attmask}_{ij} \sim \mathit{Sigmoid}({\mathbf{\lambda}}_{ij})$, follows Bernoulli distribution with parameter $\mathit{Sigmoid}({\mathbf{\lambda}}_{ij})$.  ${\attmask}_{ij}$ specifies the presence or absence of an interaction between word $i$ and word $j$ in the vocabulary $\mathbb{V}$. Hence, learning the word interaction graph reduces to learning the parameter matrix $\mathbf{\lambda}=\{ \mathbf{\lambda}_{ij} \}_{V \times V}$.  In Section~\ref{subsec:ibloss}, we  show how $\mathbf{\lambda}$ (and therefore $\attmask$) is learned  through a variational information bottleneck loss formulation (details in ~\sref{subsec:IBderivation}).

Based on the learnt interaction mask $\attmask$, each word embedding $\mathbf{x}_i$ is revised using a graph based summation from its interacting neighbors' embedding $\mathbf{x}_j, j \in \mathcal{N}(i)$:
\begin{equation}
    \e_i' =  \x_i +  \sigma\Bigg( \tfrac{1}{|\mathcal{N}(i)|}\sum_{j \in \mathcal{N}(i)} \x_j\W \Bigg)
    \label{eq:GCN_vec}
\end{equation}
 $\sigma(\cdot)$ is the ReLU non-linear activation function and $\W \in \mathbb{R}^{H \times H}$ is a weight matrix. 
We denote the resulting word representation vector as $\e'_i$. Here $j \in \mathcal{N}(i)$, and $\mathcal{N}(i)$ denotes those neighbor nodes of $\x_i$ on the graph $\attmask$ and in $\x$.  
 Eq.~(\ref{eq:GCN_vec}) is motivated by the design of Graph convolutional networks (GCNs) that were introduced to learn useful node representations that encode both node-level features and relationships between connected nodes \cite{kipf2016semi}. Differently in our work, we need to learn the graph $\attmask$, through the $\lambda$ parameter. We can compute the simultaneous update of all words in input text $\x$ together by concatenating all $\e'_i$. This gives us one matrix $\E' \in \mR^{L \times H}$, where $L$ is the length of input and $H$ is the embedding dimension of  $\x_i$.

\paragraph{Learning Word Importance Mask: $\mR$}

 This word mask aims to learn a  global attribution word mask $\R$.   Aiming for better word selection, $\R$ is designed as a learnable stochastic layer with $\R \in \{0,1\}^{V}$. Each entry in $\R$ (e.g., ${\R}_w \in\{0,1\}$ for word $w$) follows a Bernoulli distribution with parameter $p_w$. The learning reduces to learning the parameter vector $\mathbf{p}$.

During inference, for an input text $\x$, we get a binary vector ${\R}_{\x}$ from $\R$ that is of size $L$. Its $i$-th entry ${\R}_{{\x}_i}\in\{0,1\}$ is a binary random variable associated with the word token at the $i$-th position. ${\R}_{\x}$ denotes how important each word is in an input text $\x$. Then we use the following operation (a masking operation) to generate the final representation of the $i$-th word: ${\z}_i = {\R}_{\x_i} {\e'}_i$. We then feed the resulting $\bZ$ to the target model $f$. 

\subsubsection{Learning Word and Interaction Masks for  a target model $f$: }
\label{subsec:ibloss}

During training, we fix the parameters of target model $f$  and only train the \layer to get two masks.

We learn this trainable layer using the following loss objective, with the derivation of each term explained in the following section:
\begin{multline}
    \mathcal{L}(\x, f(\x),\hat{y}) = \ell_{f(\x),\hat{y}} + \beta_{sparse} \ell_{sparse}   \\ + \beta_{i} \ell{prior_{{\mR}_{\x}}} + \beta_{g} \ell_{prior_{\mA_{\x}}}
\end{multline}
First, we want to ensure that model predictions with {\layer } added are consistent with the original prediction $f(\x)$. Hence, we minimize the cross entropy loss  ${\ell}_{f(\x),\hat{y}} $ between $f(\x)$ and the newly predicted output $\hat{y}$ (when with the bottleneck layer).

Then
$\ell_{sparse}$ is the sparsity regularization on $\mA_{\x}$, $\ell{prior_{\mR}}$ is the KL divergence between $\mR$ and a random bernoulli prior. Similarly, $\ell{prior_{\mA}}$ is the KL divergence between $\mA$ and a random bernoulli prior. We provide detailed derivations in Section ~\ref{subsec:IBderivation}.

\section{Experiments}
\label{sec:experiments}
Our experiments are designed to answer the following questions: 

\begin{enumerate}
\itemsep0em 
    \item Will a test set filtered by \method find more errors from a SOTA NLP model?  
    \item Does \method achieve test adequacy faster, i.e. achieve higher coverage in fewer samples?
    \item Does \method help us compare existing testing benchmarks?
    \item Can \method help us automatically select non-redundant samples for better augmentation?
\end{enumerate}

\paragraph{Datasets and Models}
We use pretrained model BERT-base\cite{devlin2019bert} and RoBERTa-base\cite{liu2019roberta} provided by \cite{morris2020textattack} finetuned on SST-2 dataset and Quora Question Pair (QQP) dataset.  For the QQP dataset, we use the model finetuned on the MRPC dataset. We train a word level mask ($\mR$) and an interaction mask ($\mA$) for each of these settings. We use a learning rate of $1e-05$, $\beta_i=0.001$, $\beta_g=0.001$, and $\beta_s=0.001$ for all models.

We have provided the test accuracy of the target models and the models trained with masks in Table~\ref{tab:accuracy}. Note that the ground truth labels here are the {\it predictions} from the target model $f$ without the \imask layer, as our goal is to ensure fidelity of the \imask+$f$ to the target model $f$. Table~\ref{tab:accuracy} shows that training the \imask+$f$ model maintains the target model's predictions $f$ as indicated by higher accuracies.

\begin{table*}[h]
\centering
\small
\scalebox{0.8}{
\begin{tabular}{p{6cm}|rrrr|rr}
\toprule

 \multirow{2}{*}{\textbf{Test Transformation   Name}}  & \multicolumn{4}{c|}{Failure Rate ($\%$)} & \multicolumn{2}{c}{Dataset Size Reduction ($\%$)} \\
 \cline{2-7} 
 
 & D & D+\methodNoMask & D+\method & $\Delta_{D+\method}$ & D+\methodNoMask & D+\method \\
 \midrule
change names & 5.14 & ${\bf 100.00}_{16.86}$ & ${\bf 100.00}_{16.86}$ & ${94.86}_{11.73}$  & 62.84 & 62.84\\
add negative phrases & 6.80 & ${\bf 100.00}_{19.16}$ & ${\bf 99.34}_{19.16}$ & ${92.54}_{12.36}$  & 75.40 & 75.99\\
protected: race & 68.00 & ${\bf 100.00}_{73.02}$ & ${\bf 99.37}_{72.13}$ & ${31.37}_{4.13}$  & 44.33 & 43.54\\
used to,but now & 29.87 & ${\bf 53.30}_{46.27}$ & ${\bf 53.30}_{46.75}$ & ${23.43}_{16.88}$  & 84.61 & 84.61\\
protected: sexual & 86.83 & ${\bf 100.00}_{87.31}$ & ${\bf 100.00}_{88.99}$ & ${13.17}_{2.15}$  & 83.83 & 86.83\\
change locations & 8.69 & ${\bf 21.43}_{8.31}$ & ${\bf 21.43}_{8.31}$ & ${12.74}_{-0.38}$  & 86.47 & 86.93\\
change neutral words with BERT & 9.80 & ${\bf 20.00}_{13.04}$ & ${\bf 20.84}_{13.39}$ & ${11.04}_{3.59}$  & 73.40 & 73.88\\
contractions & 2.90 & ${\bf 5.52}_{3.94}$ & ${\bf 8.22}_{4.34}$ & ${5.32}_{1.44}$  & 34.80 & 34.80\\
2 typos & 11.60 & ${\bf 18.00}_{11.14}$ & ${\bf 16.00}_{10.87}$ & ${4.40}_{-0.73}$  & 10.00 & 10.00\\
change numbers & 3.20 & ${\bf 7.14}_{3.94}$ & ${\bf 7.14}_{3.29}$ & ${3.94}_{0.09}$  & 87.70 & 88.44\\
typos & 6.60 & ${\bf 10.00}_{6.45}$ & ${\bf 10.00}_{6.24}$ & ${3.40}_{-0.36}$  & 10.00 & 10.00\\
neutral words in context & 96.73 & ${\bf 100.00}_{96.91}$ & ${\bf 100.00}_{96.82}$ & ${3.27}_{0.09}$  & 21.33 & 28.50\\
protected: religion & 96.83 & ${\bf 100.00}_{93.42}$ & ${\bf 100.00}_{98.02}$ & ${3.17}_{1.19}$  & 89.67 & 89.67\\
add random urls and handles & 15.40 & ${ 14.63}_{9.17}$ & ${\bf 17.86}_{11.56}$ &${2.46}_{-3.84}$  & 87.60 & 87.60\\
simple negations: not neutral is still neutral & 97.93 & ${\bf 100.00}_{98.49}$ & ${\bf 100.00}_{98.34}$ & ${2.07}_{0.41}$  & 45.67 & 46.03\\
simple negations: not negative & 10.40 & ${\bf 12.00}_{9.56}$ & ${\bf 12.42}_{9.90}$ & ${2.02}_{-0.50}$  & 80.11 & 80.37\\
my opinion is what matters & 41.53 & ${\bf 42.49}_{36.79}$ & ${\bf 43.14}_{36.79}$ & ${1.61}_{-4.74}$  & 84.16 & 84.32\\
punctuation & 5.40 & ${\bf 6.80}_{5.37}$ & ${\bf 6.80}_{6.26}$ & ${1.40}_{0.86}$  & 10.00 & 9.11\\
Q \& A: yes & 0.40 & ${\bf 1.33}_{0.50}$ & ${\bf 1.71}_{0.49}$ & ${1.31}_{0.09}$  & 82.33 & 83.14\\
Q \& A: no & 85.20 & ${\bf 86.33}_{76.42}$ & ${\bf 86.22}_{77.16}$ & ${1.02}_{-8.04}$  & 82.34 & 83.12\\
simple negations: negative & 6.13 & ${\bf 7.33}_{5.63}$ & ${\bf 7.12}_{5.70}$ & ${0.99}_{-0.44}$  & 78.63 & 78.71\\
reducers & 0.13 & ${\bf 0.27}_{0.10}$ & ${\bf 1.03}_{0.10}$ & ${0.90}_{-0.03}$  & 67.00 & 66.77\\
intensifiers & 1.33 & ${ 1.05}_{0.31}$ & ${\bf 1.83}_{1.01}$ &${0.49}_{-0.33}$  & 66.65 & 66.96\\
\midrule
Average Improvement & - & $13.51_{1.10}$ & $13.78_{1.55}$ & $13.78_{1.55}$ & $62.99$  	& $63.57$ \\
\bottomrule
\end{tabular}}

\caption{Failure Rate(\%) obtained using BERT model on the original dataset $D$, the dataset filtered using \methodNoMask  coverage (D+\methodNoMask columns) and the dataset filtered with \method coverage (D+\method columns) from the Sentiment Test Suite. We report both the max failure rate as well as the mean in the subscript across $10$ thresholds of coverage. Rows are sorted regarding the failure rate difference between the  dataset filtered using \method and the original dataset (column $\Delta_{D+\method}$). %
}
\label{tab:failure_rate_sst2_bert}
\end{table*}

\begin{table*}[h]
\centering
\small
\scalebox{0.8}{
\begin{tabular}{p{6cm}|rrrr|rr}
\toprule

 \multirow{2}{*}{\textbf{Test  Transformation Name}}  & \multicolumn{4}{c|}{Failure Rate ($\%$)} & \multicolumn{2}{c}{Dataset Size Reduction ($\%$)} \\
 \cline{2-7} 
 & D & D+\methodNoMask & D+\method & $\Delta_{D+\method}$ & D+\methodNoMask & D+\method \\
 \midrule
Q \& A: yes & 46.20 & ${\bf 100.00}_{51.45}$ & ${\bf 100.00}_{51.45}$ & ${53.80}_{5.25}$  & 82.37 & 81.58\\
protected: race & 61.67 & ${\bf 100.00}_{66.00}$ & ${\bf 100.00}_{66.00}$ & ${38.33}_{4.34}$  & 44.33 & 44.33\\
neutral words in context & 80.87 & ${\bf 100.00}_{83.43}$ & ${\bf 100.00}_{83.31}$ & ${19.13}_{2.44}$  & 21.27 & 21.39\\
protected: religion & 73.00 & ${\bf 74.19}_{62.31}$ & ${\bf 85.71}_{63.64}$ & ${12.71}_{-9.36}$  & 89.67 & 89.67\\
protected: sexual & 91.00 & ${\bf 100.00}_{84.25}$ & ${\bf 100.00}_{84.25}$ & ${9.00}_{-6.75}$  & 83.83 & 83.83\\
simple negations: not neutral is still neutral & 91.53 & ${\bf 100.00}_{92.93}$ & ${\bf 100.00}_{92.24}$ & ${8.47}_{0.71}$  & 45.87 & 46.11\\
add negative phrases & 29.60 & ${\bf 36.46}_{23.15}$ & ${\bf 36.72}_{23.15}$ & ${7.12}_{-6.45}$  & 75.40 & 75.40\\
Q \& A: no & 57.53 & ${\bf 61.67}_{53.60}$ & ${\bf 62.41}_{53.60}$ & ${4.87}_{-3.93}$  & 82.31 & 82.31\\
simple negations: not negative & 95.40 & ${\bf 100.00}_{96.11}$ & ${\bf 100.00}_{95.92}$ & ${4.60}_{0.52}$  & 80.14 & 80.27\\
2 typos & 5.20 & ${\bf 6.40}_{5.10}$ & ${\bf 7.33}_{5.33}$ & ${2.13}_{0.13}$  & 10.00 & 10.00\\
change neutral words with BERT & 9.20 & ${\bf 11.11}_{7.81}$ & ${\bf 11.11}_{7.48}$ & ${1.91}_{-1.72}$  & 73.40 & 72.45\\
intensifiers & 1.13 & ${\bf 2.68}_{1.69}$ & ${\bf 2.68}_{1.87}$ & ${1.55}_{0.73}$  & 66.65 & 66.46\\
change names & 4.53 & ${\bf 4.88}_{2.57}$ & ${\bf 5.81}_{2.37}$ & ${1.28}_{-2.16}$  & 62.84 & 62.84\\
punctuation & 4.80 & ${\bf 6.00}_{4.03}$ & ${\bf 6.00}_{3.96}$ & ${1.20}_{-0.84}$  & 10.00 & 10.00\\
change locations & 6.16 & ${\bf 7.32}_{5.28}$ & ${\bf 7.32}_{4.44}$ & ${1.16}_{-1.72}$  & 86.47 & 87.43\\
typos & 3.00 & ${\bf 3.60}_{1.90}$ & ${\bf 4.00}_{2.39}$ & ${1.00}_{-0.61}$  & 10.00 & 10.00\\
simple negations: negative & 1.33 & ${\bf 2.00}_{1.07}$ & ${\bf 1.99}_{1.20}$ & ${0.65}_{-0.13}$  & 78.63 & 78.68\\
used to,but now & 52.73 & ${\bf 53.30}_{46.27}$ & ${\bf 53.30}_{45.58}$ & ${0.56}_{-7.16}$  & 84.61 & 83.74\\
my opinion is what matters & 56.47 & ${\bf 59.00}_{50.32}$ & ${\bf 56.86}_{49.87}$ & ${0.39}_{-6.60}$  & 84.20 & 84.20\\
contractions & 1.00 & ${\bf 1.37}_{0.74}$ & ${\bf 1.37}_{0.63}$ & ${0.37}_{-0.37}$  & 34.80 & 34.80\\
reducers & 0.40 & ${ 0.34}_{0.14}$ & ${\bf 0.57}_{0.27}$ &${0.17}_{-0.13}$  & 67.05 & 66.35\\
change numbers & {\bf 2.50} & ${1.63}_{0.48}$ & ${ 2.41}_{0.72}$ & ${-0.09}_{-1.78}$  & 87.70 & 87.70\\
add random urls and handles & {\bf 11.40} & ${8.33}_{5.22}$ & ${ 8.65}_{4.55}$ & ${-2.75}_{-6.85}$  & 87.60 & 88.18\\
\midrule
Average Improvement & - & $6.68_{-1.10}$ & $7.29_{-1.85}$ & $7.29_{-1.85}$ & $63.01$  	& $62.95$ \\
\bottomrule
\end{tabular}}

\caption{Failure Rate(\%) obtained using RoBERTa model on the original dataset $D$, the dataset filtered using \methodNoMask  coverage (D+\methodNoMask columns) and the dataset filtered with \method coverage (D+\method columns) from the Sentiment Test Suite. We report both the max failure rate as well as the mean in the subscript across $10$ thresholds of coverage. Rows are sorted regarding the failure rate difference between the  dataset filtered using \method and the original dataset (column $\Delta_{D+\method}$). 
}
\label{tab:failure_rate_sst2_rob}
\end{table*}

\begin{figure*}[htb]
  \centering
  \includegraphics[width=.3\textwidth]{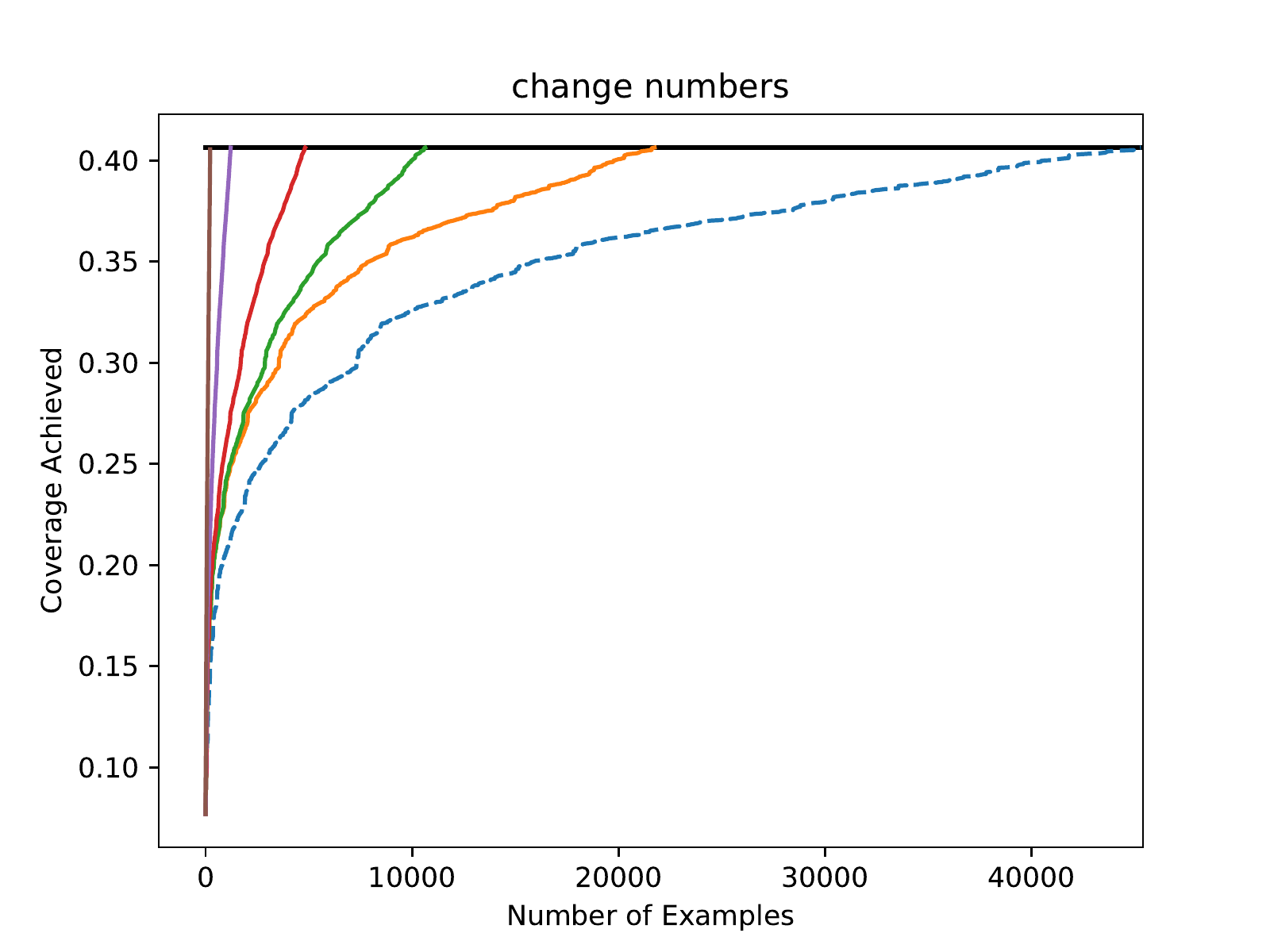}
  \includegraphics[width=.3\textwidth]{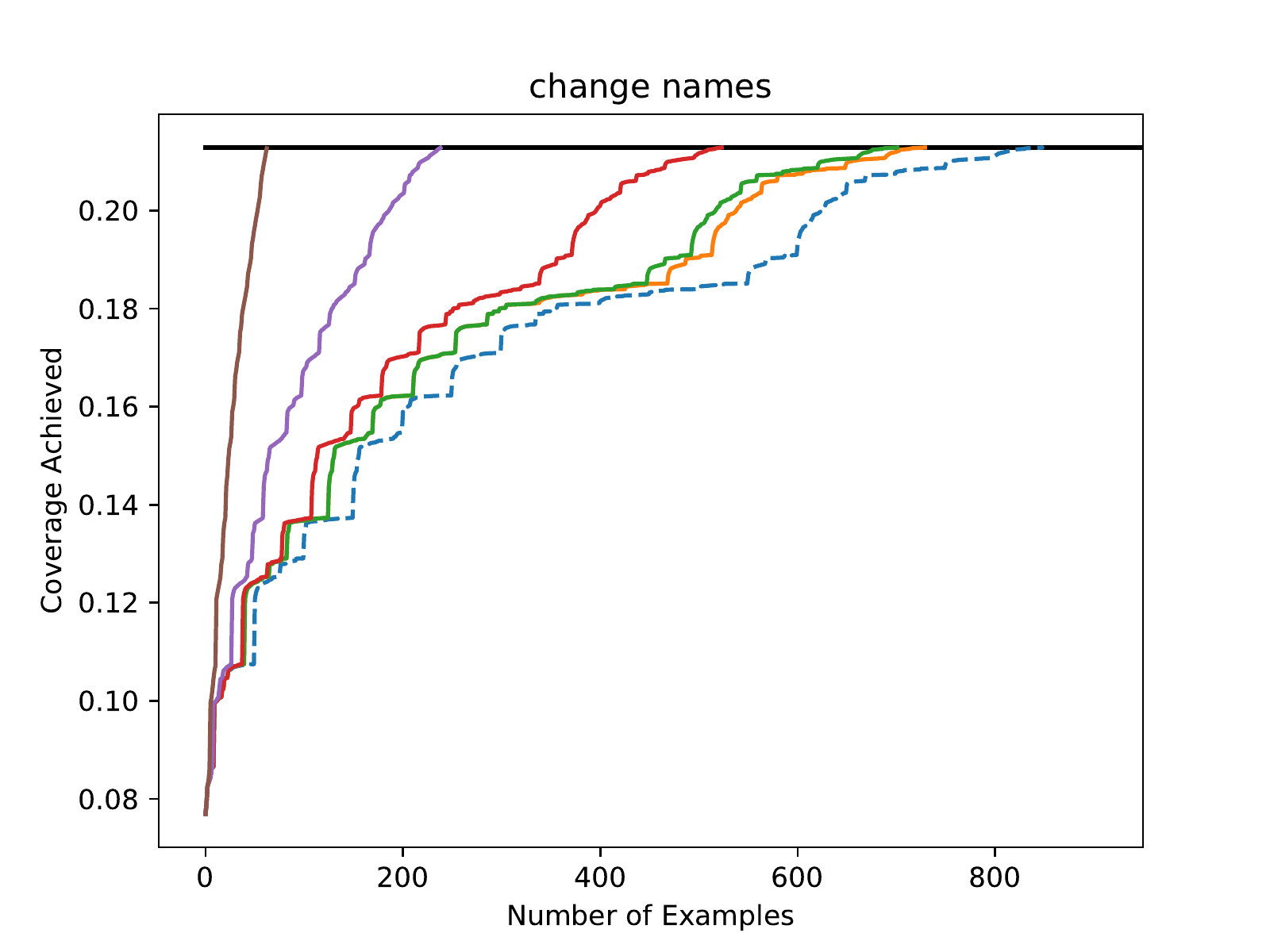}
 \includegraphics[width=.3\textwidth]{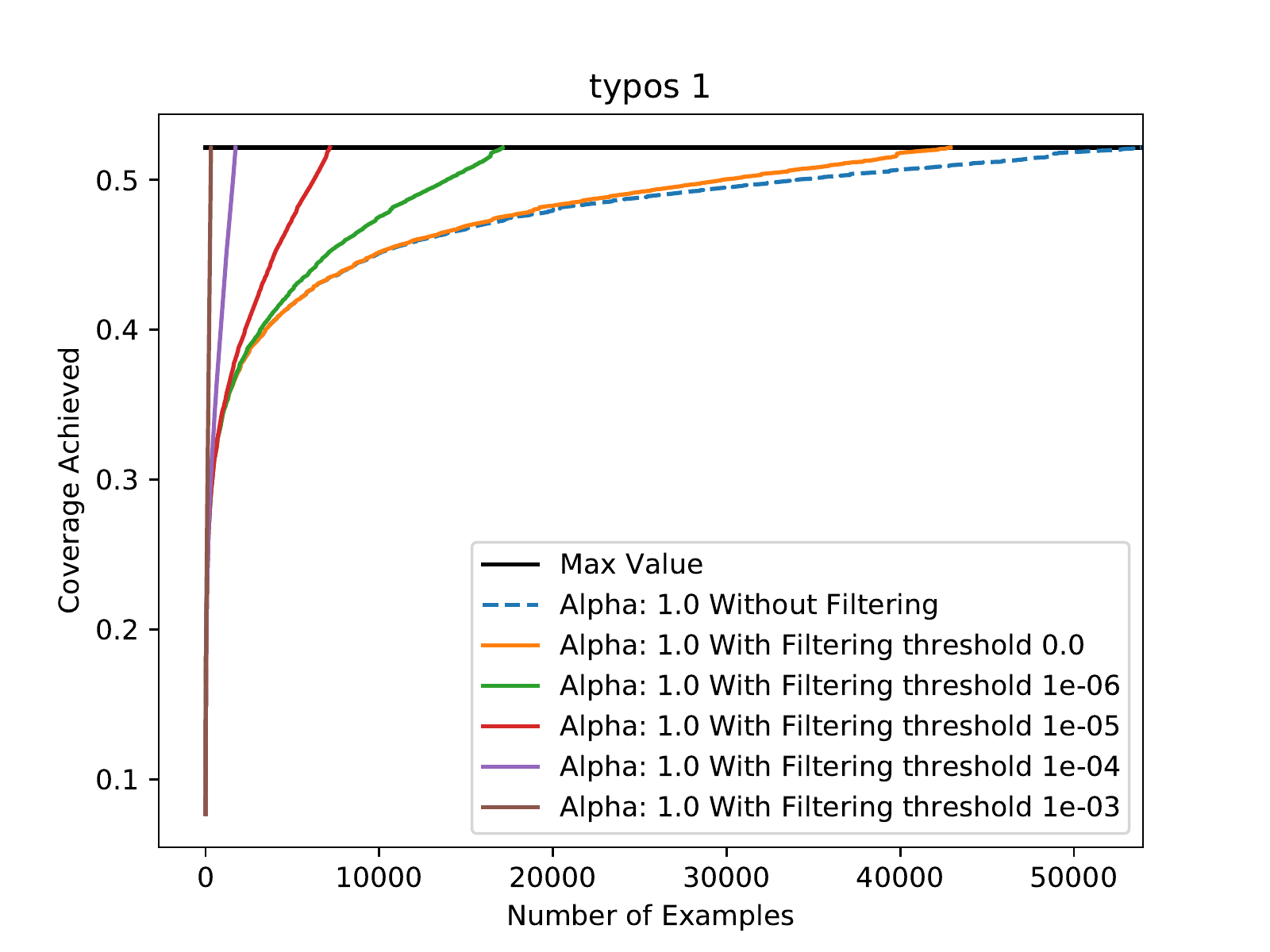}
  \includegraphics[width=.3\textwidth]{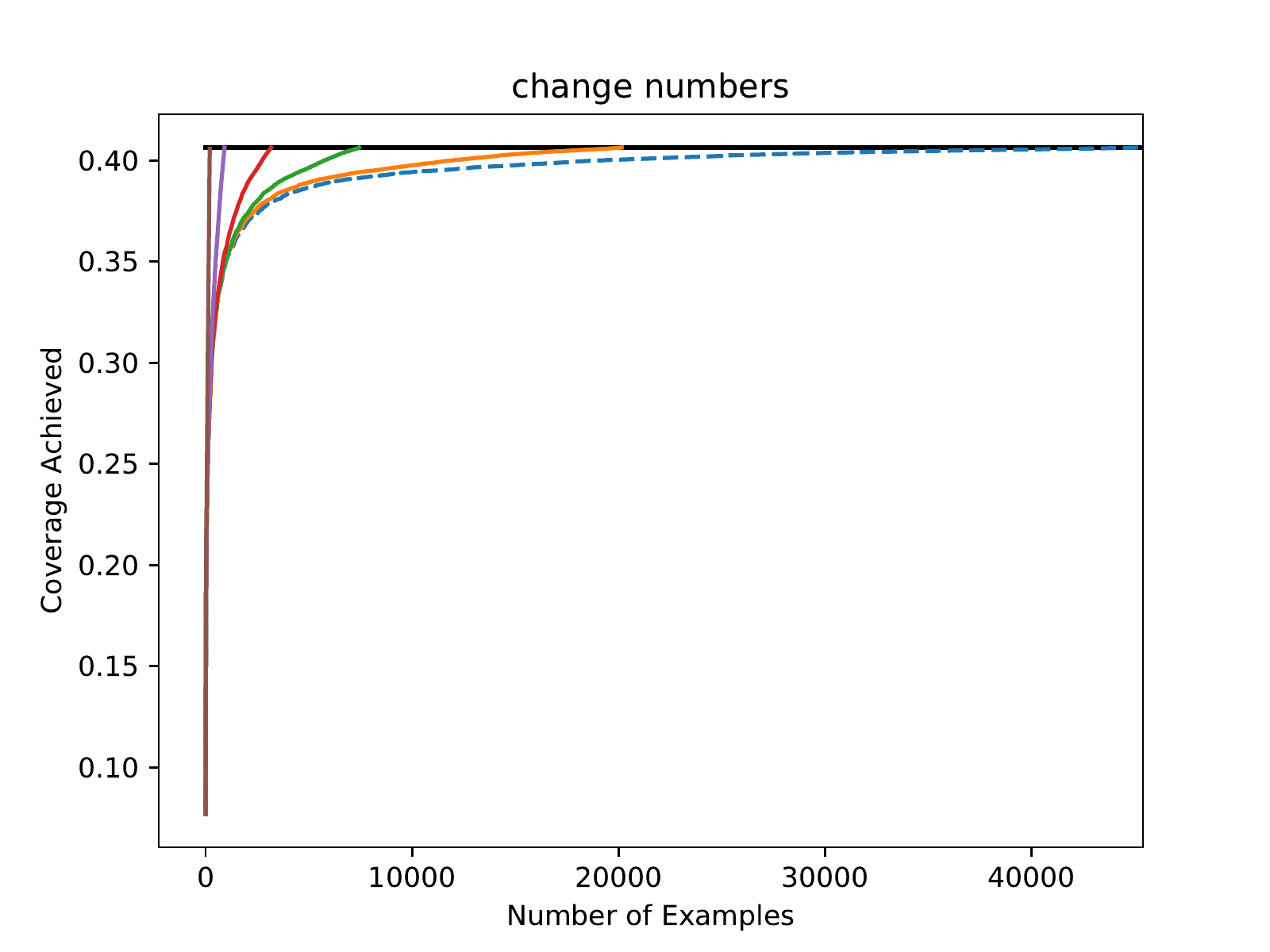}
  \includegraphics[width=.3\textwidth]{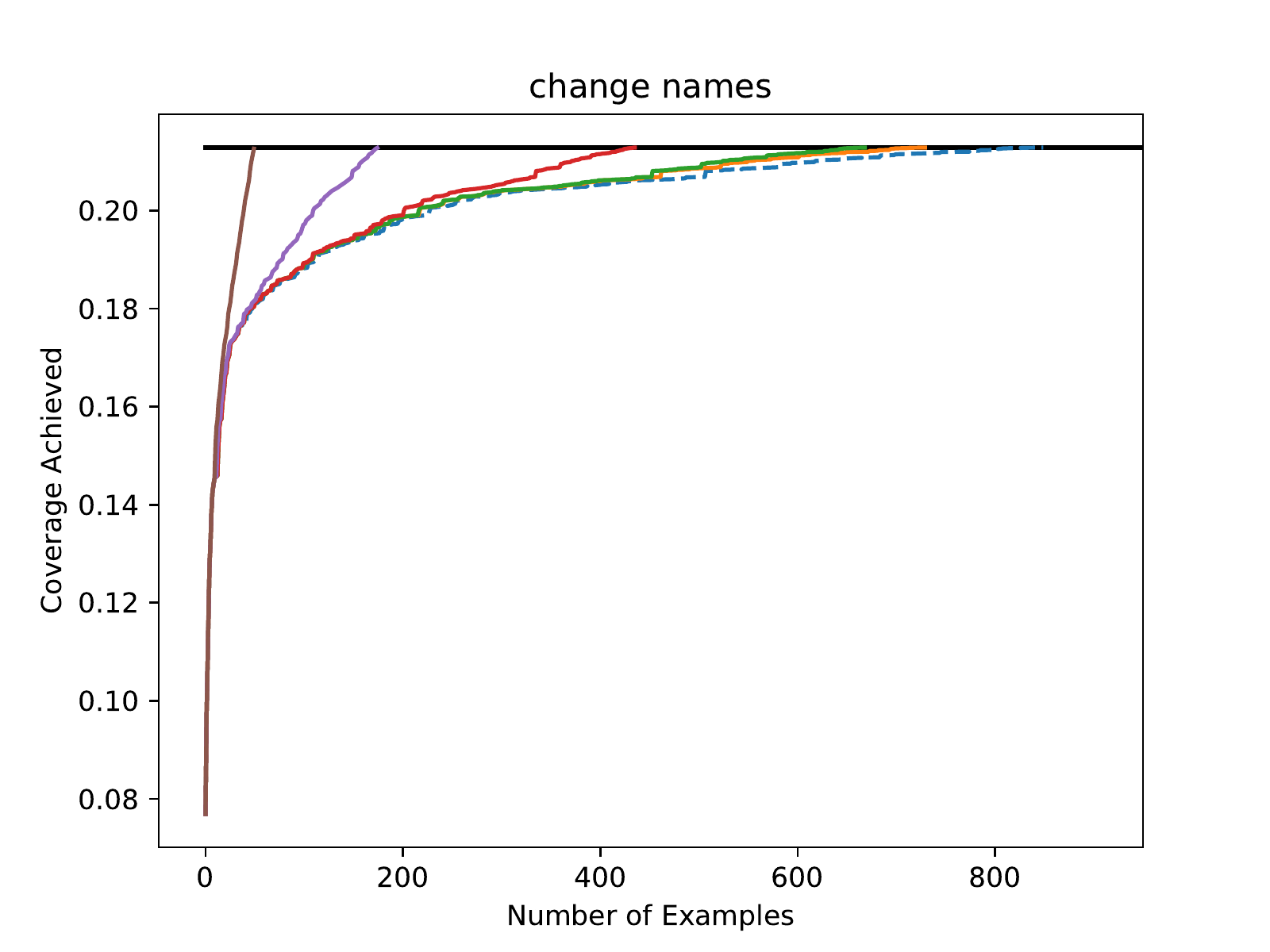}
 \includegraphics[width=.3\textwidth]{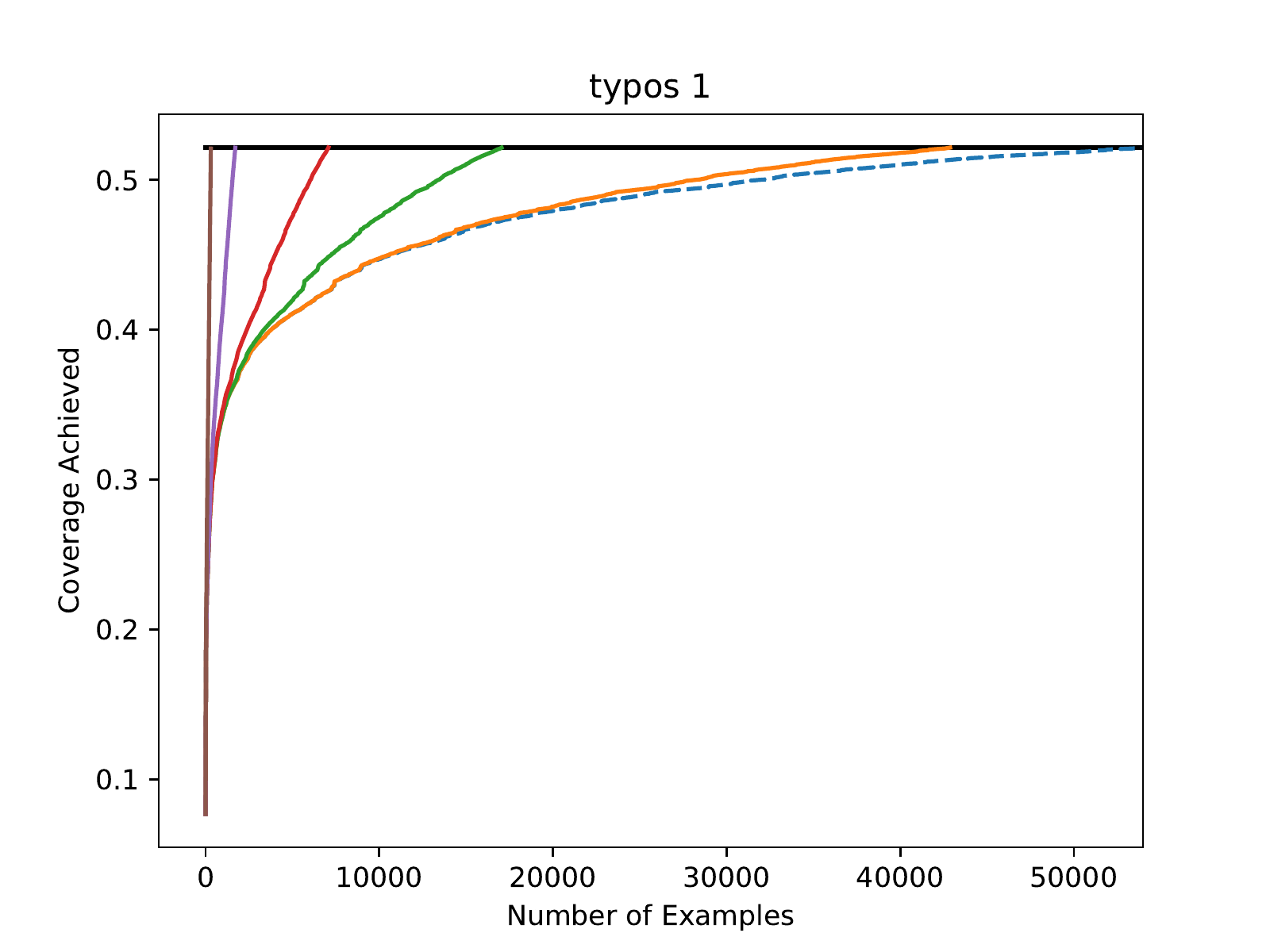}

  \caption{\method is able to achieve higher coverage with a fewer number of samples than without coverage based filtering. For this experiment we use the RoBERTa model. In the top row, we do not shuffle the examples and the bottom row with shuffling. Even with a threshold of $0.0$, we are able to significantly reduce the number of samples that achieve the same coverage as the unfiltered set: we are able to achieve an average  reduction (across transformations) of 28.83\%,	54.06\%	,	71.48\%	,	88.67\%	and	97.17\%	 for $\{0.0, 1e-04, 1e-03, 1e-02, 1e-01\}$ thresholds respectively.}
  \label{fig:coverage_saturation}
\end{figure*}
\subsection{Experiment 1: Removing Redundant Test Inputs during Model Testing}
\paragraph{Motivation} \Checklist\cite{ribeiro2020accuracy} provides a method to generate a large number of test cases corresponding to a target template. It introduces different transformations that can be used to generate samples to check for a desired behavior/functionality. For example, to check for a model's behavior w.r.t typos in input texts, it generates examples with typos and queries the target model. \Checklist then compares failure rates across models for the generated examples to identify failure modes. However, it does not provide a method to quantify the number of tests that need to be generated or to determine which examples provide the most utility in terms of fault detection. Such a blind generation strategy may suffer from sampling bias and give a false notion of a model's failure modes. 

\paragraph{Setup} We evaluate \method's ability to redundant test samples out of the generated tests for a target model. The different transformations used are summarized in Column I of \tref{tab:failure_rate_sst2_bert}, ~\tref{tab:failure_rate_sst2_rob},   and ~\tref{tab:failure_rate_qqp}. We measure failure rate on an initial test set of size $N=1500$. We then filter the generated tests based on our \method coverage criteria:  if adding a test example into a test suite does not lead to an increase in its coverage, it is discarded. We then measure failure rate of the new filtered test set.

\paragraph{Results} In \tref{tab:failure_rate_sst2_bert}, we observe that \method can select more failure cases for the BERT model across all 23 transformation on SST-2 dataset. On average, both \method and \methodNoMask can help reduce more than 60\% of the test suite size, and \method achieves a slight advantage over \methodNoMask.  In~\tref{tab:failure_rate_sst2_rob}, when using RoBERTa model, \method based filtering wins over the original dataset  in 21 of 23 cases. The average improvement of \method regarding error detection is 7.29\% on  RoBERTa model and 13.78\% on BERT model.  We include similar results on QQP dataset in \tref{tab:failure_rate_qqp}.

\subsection{Experiment 2: Achieving Higher Dataset Coverage with Fewer Data Points}

\paragraph{Motivation} We revisit the question: given a test generation strategy, does adding more test samples necessarily add more information?  In this set of experiments, we appeal to the software engineering notion of “coverage" as a metric of test adequacy. We show that we can reach a target level of test adequacy faster, i.e. a higher coverage, hence achieving more rigorous behavior testing, with fewer test examples, by using coverage as an indicator of redundant test samples.

\paragraph{Setup:} We use the training set as seed examples and generate samples using transformations used in the previous experiment listed in Table ~\ref{tab:failure_rate_sst2_bert}. Similar to the previous set of examples, we disregard an example if the increase in its coverage is below $threshold$. We vary these $threshold$ values  $\in \{1e-04, 1e-03, 1e-02, 1e-01, 0.0\}$. Higher the $threshold$, more number of examples get filtered out.  

\paragraph{Results:} In Figure~\ref{fig:coverage_saturation}, we show that using our coverage guided filtering strategy, we are able to achieve coverage with a fewer number of samples than without coverage based filtering. 

Even with a threshold of $0.0$, we are able to significantly reduce the number of samples that achieve the same coverage as the unfiltered set: we are able to achieve an average reduction across transformations (higher the better) of 71.17\%, 	45.94\%, 	28.52\%, 11.33\% and	2.83\% for $\{0.0, 1e-04, 1e-03, 1e-02, 1e-01\}$ thresholds respectively.

\begin{table}[h]
	\centering

    \begin{tabular}{cc}
		\toprule
		Test Set & \method \\
		\midrule
		 A1 & 0.175 \\
		 A1 + A2 & 0.182 \\
		 A1 + A2 + A3 & 0.185 \\ \midrule
		 A2 & 0.179  \\
		 A3 & 0.181 \\
			\bottomrule
	\end{tabular}
	\caption{\method Values on the Dynabench test sets}
	\label{tab:dyna}

\end{table}

\subsection{Experiment 3: \method as a Metric to Evaluate Testing Benchmarks}
\paragraph{Motivation: } In this set of experiments, we utilize coverage as a test/benchmark dataset evaluation measure. Static test suites, such as the GLUE benchmark, saturate and become obsolete as models become more advanced. To mitigate the saturation of static benchmarks with model advancement, \cite{kiela2021dynabench} introduced Dynabench, a dynamic benchmark for Natural Language Inference(NLI). Dynabench introduced a novel human-and-model-in-the-loop dataset,
consisting of three rounds that progressively increase in difficulty and complexity.  This results in three sets of training, validation and test datasets, with increasing complexity testing datasets. We use \method as an additional validation measure for the datasets. 

\paragraph{Setup:} We test the ROBERTA-Large model provided by Dynabench trained on training data from all three rounds of the benchmark. We use $10$ as the number of bins and $\lambda=1.0$. 

\paragraph{Results: } We measure coverage achieved by each of the test sets individually as well as in combination. We have summarized the results in Table~\ref{tab:dyna}. The test sets indeed provide more novel test inputs to the model as indicated by the increasing coverage as the test sets from each split are taken into consideration. The low values arise from a large architecture, (24-layer, 1024-hidden, 16-heads) that is potentially still unexplored with $1000$ samples from each test set. 

\subsection{Experiment 4: Coverage Guided Augmentation}
\paragraph{Motivation:} Data augmentation refers to strategies for increasing the diversity of training examples without
explicitly collecting new data. This is usually achieved by transforming training examples using a transformation. A number of automated approaches have been proposed to automatically select these transformations including like \cite{xie2019unsupervised}. Since computing \method does not require retraining, and the input selection can indicate the usefulness of a new sample, we  propose to use \method to select transformed samples, in order to add them into the training set for improving test accuracy.

\paragraph{Setup: } In this set of experiments, we focus on using coverage to guide generation of augmented samples. We propose  a greedy search algorithm to coverage as guide to generate a new training set with selected augmentations. The procedure is described in Algorithm~\ref{alg1} and is motivated by a similar procedure from \cite{tian2018deeptest}. This is a coverage-guided greedy search technique for efficiently
finding combinations of transformations that result in higher
coverage. 
We use transformations described in ~\sref{subsec:moreExpt} and BERT model pretrained on the datasets.

We then add the coverage selected   samples into the training set and retrain a target model. Using BERT model as base, Table~\ref{tab:aug} shows the test accuracy, when with or without adding the selected samples  into the training set. We also show  the size of the augmentation set. Our results show that using \method to guide data augmentation can improve test accuracy in both SST-2 and QQP.

\begin{table}[bth]
\centering
\small
\scalebox{0.8}{	\begin{tabular}{c|c|c|c} \toprule
Dataset & Coverage  & Test  & Size of  \\ 
& Threshold & Accuracy  & Augmented Set\\ \midrule
\multirow{3}{*}{SST-2} & Baseline & 90.22 & 0 \\ 

& Random       &     90.45           & 6541                         \\
& \method         & 90.41                   & 6541                         \\ \midrule

\multirow{3}{*}{QQP} & Baseline & 90.91  & 0 \\

& Random    &     90.96           & 14005                        \\ 
& \method      & 91.03                & 14005                        \\ \bottomrule
\end{tabular}}
\caption{ The test accuracy after adding the augmented set generated using coverage guidance to the training set on SST-2 and QQP dataset.  }
\label{tab:aug}
\end{table}

\section{Related Work}

Our work connects to a few topics in the literature. 

\paragraph{Testing for Natural Language Processing}
Recent literature has shown that deep learning models often exhibit unexpectedly poor behavior when deployed ``in the wild". This has led to a growing interest in testing NLP models. The pioneering work in this domain is \Checklist\cite{ribeiro2020accuracy}, that provides a behavioral testing template for deep NLP models. A different paradigm is proposing more thorough and extensive evaluation sets. For example, \cite{kiela2021dynabench} and \cite{koh2021wilds} proposed new test sets 
reflecting distribution shifts that naturally arise in real-world language applications. On a similar line, \citep{belinkov2019analysis, naik2018stress} introduced challenge set based testing. Another line of work has focused on perturbation techniques for evaluating models, such as logical consistency \citep{ribeiro2019red}, robustness to noise \citep{belinkov2017synthetic}, name changes \citep{prabhakaran2019perturbation}, and adversaries \citep{ribeiro2018semantically}.

\paragraph{Subset Selection}
Our \method can be used as a guide for filtering test inputs, and hence is a data selection approach. Previous work have looked at finding representative samples from training and/or interpretation perspectives. For example, submodular optimization from \cite{lin2009select, lin2010application} provides a framework for selecting examples that minimize redundancy with each other to select representative subsets from large data sets. These methods are part of the ``training the model" stage, targeting to achieve higher accuracy with fewer  training samples. Moreover, Influence Functions from \cite{koh2020understanding} provide a strategy to interpret black box models by discovering important representative training samples. The influence function can  explain and attribute a model's prediction  back to its training samples. Differently, \method is a test suite evaluation approach. 

\section{Conclusion}
This paper proposes \method to perform white-box coverage-based behavior testing on NLP models. We design \method to consider Transformer models' properties, focusing on essential words and important word combinations. Filtering test sets using the \method helps us reduce the test suite size and improve error detection rates. We also demonstrate that \method serves as a practical criterion for evaluating the quality of test sets. It can also help generate augmented training data to improve the model's generalization.

\bibliography{refcoverage,refdnntesting,refmltesting,refsoftwaretesting}

\appendix
\section{Appendix}

\subsection{Deriving How Two Masks are Used}

To learn these global masks, we update each preloaded word embedding $\x_i \forall i \in \{1, \dots, L\}$ using  embeddings from words that interact with $\x_i$ as defined by the learnt interaction matrix ${{\attmask}}_{\x}$. Specifically, to get interaction-based word composition, we use the following formulation:

\begin{equation}
\e_i' = (\e_i + g({{\attmask}}_{[\x_i,:]},\mathbf{E}))
\label{eq:imask}
\end{equation}
Here, $\mathbf{e'}_i$ is the updated word embedding for token $\x_i$ after taking into account its interaction scores with other words in the sentence $\mathbf{E} = [\e_1, \dots, \e_L]$. This is motivated from the message passing paradigm from \cite{kipf2016semi}, where we treat each word in a sentence as a node in a graph. Using Equation~\ref{eq:imask}, we effectively augment a word's embedding using information from words it interacts with.  Note that we normalize ${{{\attmask}}}_{\x}$, using $\mathbf{D}^{-1/2}{{{\attmask}}}_{\x}\mathbf{D}^{-1/2}$, where $\mathbf{D}$ is the diagonal node degree matrix for ${{\attmask}}_{\x}$.  
$g({\attmask}_{\x},\e_{\{j\}})\  \forall j \in \{1, \dots, L\}$ is the aggregation function.  Equation~\ref{eq:imask} formulation represents words and their local sentence level neighborhoods' aggregated embeddings. Specifically, we use $g({{\attmask}}_{\x},\mathbf{E}_{\x}) = h({{\attmask}}_{\x} \mathbf{E}_{\x})$. Here, $h$ is a non-linearity function, we use the ReLU non linearity. Simplifying our interaction based aggregation, if two words $x_i, x_j$ are related in a sentence, we represent each word using $\e'_i =  (\e_i + \sigma(a_{ij}(\e_i+\e_j)))$. Similarly,  $\e'_j =  (\e_j + \sigma(a_{ji}(\e_i+\e_j)))$. %
Further, to select words based on interactions,  we add a word level mask ${\wmask}$ after the word
embeddings, where ${\wmask} = [{\wmask}_{\x_1}
, \dots, {\wmask}_{\x_L}
]$.

 $\mathbf{z}_i = {\wmask}_i* \e'_i$, where ${\wmask}_i \in \{0,1\}$  is a binary
random variable. 
$\bZ = [\z_1, \dots, \z_L]$ represents word level embeddings input into a model for a specific input sentence after passing through the bottleneck layer.

\subsection{Deriving the Loss}
\label{subsec:IBderivation}
We introduce a bottleneck loss:
  \begin{equation}
        \ell_{IB} = max_{\mathbf{Z}} I(\mathbf{Z};\mathbf{Y}) - \beta I(\mathbf{Z}; \mathbf{X}) 
        \label{eq:ib_main_objective}
    \end{equation}  
Given $X$, we assume $\mathbf{E'}$ and ${\wmask}$ are independent of each other. We write $q(\bZ|\X) =q({\wmask}|\X)q(\mathbf{E'}|\X)$. From Equation~\ref{eq:imask}, 
$\mathbf{e'}_i = \mathbf{e}_i + \text{ReLU}({\attmask}_{\x} \mathbf{E}_{[1, \dots, L]})$. 
$q(\mathbf{E'}|\X)$ can be written as $q({\attmask}_{\x} | \X)$.

The lower bound to be maximized is: 
\begin{equation}
\begin{aligned}
  \mathcal{L}= \mathbb{E}_{q(\mathbf{Z}|\x^m)} log (p(\mathbf{y}^m| {\wmask}, {\attmask}, \x^m)) \\
   - \beta_i KL(q({\wmask} | \x^m) || p_{r0}({\wmask})) \\
   - \beta_g KL(q({\attmask} | \x^m) || p_{a0}({\attmask}))
   \end{aligned}
\end{equation}
We use the bernoulli distribution prior (a non informative prior) for each word-pair interaction $q_{\phi}[{\attmask}_{x_i, x_j}|\x_i,\x_j ]$. $p_{a0}({\attmask}_{x})=\prod_{i=1}^L\prod_{j=1}^L p_{a0}({\attmask}_{\x_i, \x_j})$, hence $p_{a0}({\attmask}_{x_i, x_j})= Bernoulli(0.5)$.  This leads to:
\begin{equation}
\begin{aligned}
    \beta_g KL(q({\attmask}_{\x} | \x^m) || p_{a0}({\attmask})) = \\
    -\beta_g H_{\mathbf{q}} ({\attmask}_{\x}| \x^m) 
\end{aligned}
\end{equation}
Similarly, we use the same bernoulli distribution prior for the word mask, $p_{r0}({\wmask}) = \prod_{i=1}^L p_{r0}({\wmask}_{\x_i})$, and $p_{r0}({\wmask}_{\x_i})=Bernoulli(0.5)$:
\begin{equation}
\begin{aligned}
    \beta_i KL(q({\wmask}_{\x} | \x^m) || p_{a0}({\wmask})) = \\
    - \beta_r H_{\mathbf{q}} ({\wmask}_{\x}| \x^m)
\end{aligned}
\end{equation}

We also add a sparsity regularization on ${\attmask}_{\x}$ to encourage learning of sparse interactions. Finally, we have the following loss function:
\begin{equation}
\begin{split}
    & L = - (E_{\x} p(\y|\x^m, {\attmask}, {\wmask}) + \\
    & \beta_i H_q({\wmask}|\x^m) \\ + 
    & \beta_{g} H_{\mathbf{q}}({\attmask}_{\x}| \x^m)) +\\
    & \beta_{sparse} ||{\attmask}_{\x}||_1 
    \end{split}
\end{equation}

As ${\attmask}$ is a binary graph sampled from a Bernoulli distribution with parameter $\mathbf{\gamma}$, to train the learnt parameter $\mathbf{\gamma}$, we use the Gumbel-Softmax\cite{jang2016categorical} trick to differentiate through the sampling layer. To learn the word mask ${\wmask}$, we use the amortized variational inference\cite{rezende2015variational}. We use a single-layer feedforward neural network as the inference network $q_{\phi(R_{x_t})| \x_t}$, whose parameters are optimized with the model parameters
during training. We use Gumbel-Softmax for training with discrete word mask.

\begin{minipage}{.95\linewidth}

\removelatexerror
\begin{algorithm}[H]\small
\caption{Coverage Guided Greedy Search to generate Augmented Set $G$\label{alg1} }
\SetAlgoLined
\KwResult{Test Set $G$}
 Set of Transformations $T$, Initial Seed Test set $S$\;
 \While{S is not empty}{
 text0 = S.pop()\;
 cov0 = cov(text0)\;%
 text = text0\;
 Tqueue = $\phi$\;
 iter = 0\;
  \While{$iter \leq maxIter$}{
  
  \eIf{Tqueue is not empty}{
   T1 = Tqueue.dequeue()\;
   
   }{
  T1 = RandomFrom($T$)\;
  }
  T2 = RandomFrom($T$)\;
  text1 = ApplyTransform(text, T1, T2)\;
   \eIf{covInc(text1, cov0) and CosineSim(text1, text)} {
   text = text1\;
   Tqueue.enqueue(T1)\;
   Tqueue.enqueue(T2)\;
   G.append(text)\;
   break\;
   }{
  iter += 1\;
  }
 }
 }
 \end{algorithm}
\end{minipage}

\begin{table}[tb!h]
    \centering
    \begin{tabular}{c|c|c} \toprule
        Model & Dataset & Test Accuracy  \\ \midrule
      \multirow{2}{*}{BERT}   & SST-2 & 99.31 \\
      & QQP & 99.77 \\
       \multirow{2}{*}{RoBERTa}   & SST-2 & 97.36 \\
      & QQP & 99.66 \\ \bottomrule

    \end{tabular}
    \caption{Test accuracy (in \%) of models trained with \imask layer. Note that the ground truth labels here are the {\it predictions} from the target model $f$ without the \imask layer, as our goal is to ensure fidelity of the \imask+$f$ to the target model $f$. The original models' accuracies are summarized in Table ~\ref{tab:aug}.} %
    \label{tab:accuracy}
\end{table}
\subsection{More Details and Results on Experiments}
\begin{figure}
    \centering
    \includegraphics[width=\columnwidth]{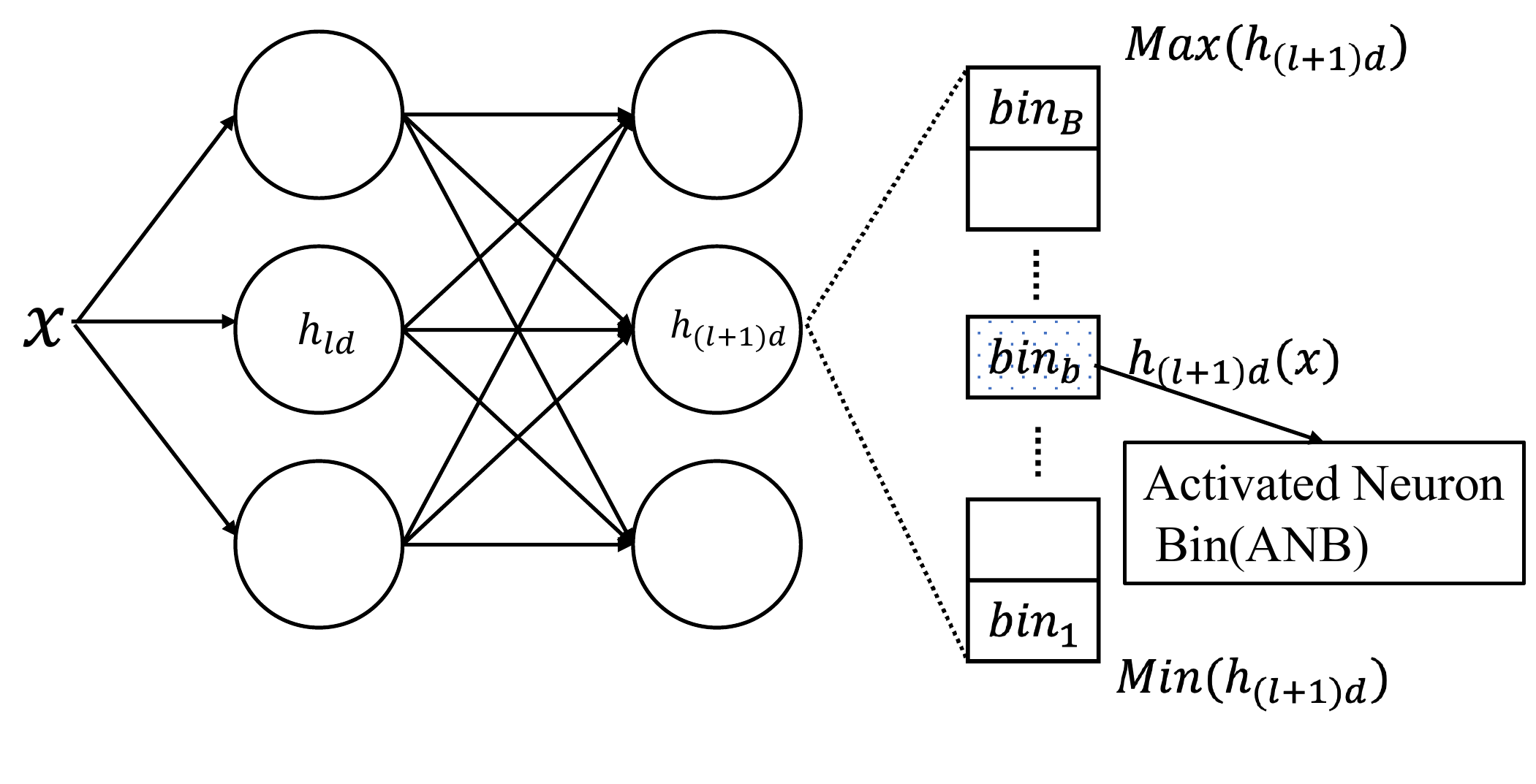}
    \caption{A schematic of k-multisection coverage in a DNN model.}
    \label{fig:anb}
\end{figure}
\label{subsec:moreExpt}
For Experiment 4.4 Coverage Guided Augmention, the set of transformations we consider are : RandomSynonymInsertion, WordSwapEmbedding, WordSwapChangeLocation, WordSwapChangeName, WordSwapChangeNumber, 
WordSwapContract, WordSwapExtend, WordSwapHomoglyphSwap, WordSwapMaskedLM, WordSwapQWERTY, WordSwapNeighboringCharacterSwap, WordSwapRandomCharacterDeletion, WordSwapRandomCharacterInsertion,  WordSwapRandomCharacterSubstitution ,RandomSwap, and WordSwapWordNet.

\begin{table*}[h]
\centering
\small
\scalebox{0.8}{
\begin{tabular}{p{6cm}|rrrr|rr}
\toprule

 \multirow{2}{*}{\textbf{Test  Transformation Name}}  & \multicolumn{4}{c|}{Failure Rate ($\%$)} & \multicolumn{2}{c}{Dataset Size Reduction ($\%$)} \\
 \cline{2-7} 
 & D & D+\methodNoMask & D+\method & $\Delta_{D+\method}$ & D+\methodNoMask & D+\method \\
 \midrule
Change first name in one of the questions & 63.00 & ${\bf 100.00}_{59.40}$ & ${\bf 100.00}_{64.49}$ & ${37.00}_{1.49}$  & 98.20 & 98.20\\
add one typo & 19.40 & ${\bf 28.57}_{23.17}$ & ${\bf 29.41}_{23.52}$ & ${10.01}_{4.12}$  & 88.00 & 88.00\\
Product of paraphrases(q1) * paraphrases(q2) & 95.00 & ${\bf 100.00}_{100.00}$ & ${\bf 100.00}_{100.00}$ & ${5.00}_{5.00}$  & 99.00 & 99.00\\
Replace synonyms in real pairs & 8.37 & ${\bf 13.33}_{7.89}$ & ${\bf 12.50}_{8.81}$ & ${4.13}_{0.44}$  & 73.31 & 73.31\\
Symmetry: f(a, b) = f(b, a) & 6.00 & ${\bf 8.33}_{4.83}$ & ${\bf 10.00}_{4.61}$ & ${4.00}_{-1.39}$  & 82.00 & 82.00\\
Testing implications & 15.07 & ${\bf 15.25}_{7.90}$ & ${\bf 15.25}_{7.46}$ & ${0.19}_{-7.60}$  & 99.29 & 99.29\\
same adjectives, different people v3 & {\bf 100.00} & ${\bf 100.00}_{100.00}$ & ${\bf  100.00}_{100.00}$ & ${0.00}_{0.00}$  & 81.82 & 81.82\\
same adjectives, different people & {\bf 100.00} & ${\bf 100.00}_{100.00}$ & ${ \bf  100.00}_{100.00}$ & ${0.00}_{0.00}$  & 81.48 & 81.48\\
Change same location in both questions & {\bf 5.00} & ${0.00}_{0.00}$ & ${ 0.00}_{0.00}$ & ${-5.00}_{-5.00}$  & 96.60 & 96.60\\
\midrule
Average Improvement & - & $5.96_{-0.96}$ & $6.15_{-0.33}$ & $6.15_{-0.33}$ & $88.86$  	& $88.86$ \\
\bottomrule
\end{tabular}}

\caption{Failure Rate(\%) obtained using BERT model on the original dataset $D$, the dataset filtered using \methodNoMask  coverage (D+\methodNoMask columns) and the dataset filtered with \method coverage (D+\method columns) from the QQP Suite. We report both the max failure rate as well as the mean in the subscript across $10$ thresholds of coverage. Rows are sorted regarding the failure rate difference between the  dataset filtered using \method and the original dataset (column $\Delta_{D+\method}$). We use $200$ samples in this case.  
}
\label{tab:failure_rate_qqp}
\end{table*}

\end{document}